\documentclass[10pt,twocolumn,letterpaper]{ieeeconf}

\usepackage{times}
\usepackage{epsfig}
\usepackage{graphicx}
\usepackage{amsmath}
\usepackage{amssymb}
\usepackage{mathtools}
\usepackage{subcaption}
\usepackage{multirow}
\usepackage{booktabs}

\begin{document}

\title{Decompose to manipulate: Manipulable Object Synthesis in 3D Medical Images with Structured Image Decomposition}

\author{Siqi Liu$^{\dagger}$, Eli Gibson$^{\dagger}$, Sasa Grbic$^{\dagger}$, Zhoubing Xu$^{\dagger}$, Arnaud Arindra Adiyoso Setio$^{\bullet}$, \\
Jie Yang$^{\diamond\star}$\thanks{$\star$ This work was conducted when Jie Yang was an intern AI scientist at Siemens Healthineers.}, Bogdan Georgescu$^{\dagger}$, Dorin Comaniciu$^{\dagger}$ \\
$^{\dagger}$Digital Services, Digital Technology \& Innovation, Siemens Healthineers, Princeton, NJ, USA\\
$^{\bullet}$ Digital Services, Digital Technology \& Innovation, Siemens Healthineers, Erlangen, Germany\\
$^{\diamond}$ Department of Biomedical Engineering, Columbia University, New York, NY, USA 
{\tt\small siqi.liu@siemens-healthineers.com}
}

\maketitle

\begin{abstract}
The performance of medical image analysis systems is constrained by the quantity of high-quality image annotations.  Such systems require data to be annotated by experts with years of training, especially when diagnostic decisions are involved. Such datasets are thus hard to scale up. 
In this context, it is hard for supervised learning systems to generalize to the cases that are rare in the training set but would be present in real-world clinical practices.
We believe that the synthetic image samples generated by a system trained on the real data can be useful for improving the supervised learning tasks in the medical image analysis applications. 
Allowing the image synthesis to be manipulable could help synthetic images provide complementary information to the training data rather than simply duplicating the real-data manifold.
In this paper, we propose a framework for synthesizing 3D objects, such as pulmonary nodules, in 3D medical images with manipulable properties. 
The manipulation is enabled by decomposing of the object of interests into its segmentation mask and a 1D vector containing the residual information.
The synthetic object is refined and blended into the image context with two adversarial discriminators.
We evaluate the proposed framework on lung nodules in 3D chest CT images and show that the proposed framework could generate realistic nodules with manipulable shapes, textures and locations, etc.
By sampling from both the synthetic nodules and the real nodules from 2800 3D CT volumes during the classifier training, we show the synthetic patches could improve the overall nodule detection performance by average 8.44\% competition performance metric (CPM) score.
\end{abstract}
\section{Introduction}

The performance of deep learning systems in medical image analysis applications is constrained by the quantity of high-quality image annotations. 
Large-scale datasets for training and testing are essential to reduce the variance of the trained networks in supervised learning as well as providing a reliable estimate of their long-term performance after deployment.
Most of the medical image datasets only scale from hundreds to thousands of patients acquired from few clinical imaging sites.
Different from annotating the natural image datasets, the diagnostic AI applications normally require the annotators to have years of medical training, and thus are expensive to scale due to the time and financial cost.
The distribution of such images is highly biased towards only a small portion of the global population.
Also, rare abnormalities may have too few exemplars in the training dataset to generalize well to prospective patients.

The data efficiency of such learning systems is thus essential.
It can be either improved via (1) better supervised learning algorithms, such as the network architectures, optimization algorithms, and objective functions, or (2) synthesizing images and their annotations from the manually annotated images, for example, the data augmentation techniques with simple image transformations.
We explore the latter path in this work by synthesizing objects in medical images with high fidelity while allowing their properties to be manipulable.
This manipulability is important in the context of limited medical image datasets as it allows synthesis to (1) reproduce the variability of semantically meaningful features that are observed in the clinic but not in the limited dataset and (2) over-sample realistic but challenging samples where system performance is more clinically important.

We propose the 3D manipulable object synthesis with structured object decomposition and adversarial image refining. 
We start with training a conditional variational autoencoder (cVAE) \cite{RezendeVAE,KingmaVAE} on mesh vertices to generate realistic 3D object meshes.
We then train an image decomposition network to decompose the object patch into a segmentation mask and a 1D vector containing the residual information related to the object intensity, texture, etc.
A decoder network is trained to reconstruct the object patch from the decomposed components and further blend the reconstructed object into its context.
In the last training stage, the decoder is fine-tuned by applying two adversarial discriminators to synthesize objects on patches initially without the target objects.
Different from the existing object in-painting methods in medical images, the proposed framework allows the shapes, sizes and intensities of the generated objects to be manipulated.
We evaluated the proposed framework on an example application of synthesizing lung nodules in 3D CT images and using the synthetic nodules to improve the nodule detection performance.
With a dataset of 2800 3D CT volumes in total, we show that the synthetic nodule patches could improve the lung nodule detection performance.
Our contributions can be summarized as:
(1) an object synthesis framework which can synthesize objects, such as lesions, in 3D medical images with manipulable properties at random locations;
(2) investigating the application of using synthetic patches to improve the detection of pulmonary nodules.

\section{Related Work}
\subsection{Medical Image In-painting}
Several recent studies \cite{Jin2018CT-RealisticSegmentation,Shin2018MedicalNetworks,Korkinof2018High-ResolutionNetworks,Lau2018ScarGAN:Scans,Wu2018ConditionalClassification,Frid-Adar2018GAN-basedClassification} attempted to explore the application of adversarial image synthesis in medical image analysis. 
In \cite{Jin2018CT-RealisticSegmentation}, the authors proposed to use a fully convolutional neural network to in-paint lung nodules in a masked area of a 3D chest CT image patch.
The network output is sent to an adversarial discriminator network to ensure the realism of the synthetic nodule.
The appearance of the in-painted nodules is only conditioned on the context of the in-painting area,
the effect on the data augmentation might be highly limited in many applications if the diversity the synthetic objects cannot be controlled.
In a similar work \cite{Wu2018ConditionalClassification}, the generated objects are conditioned by the segmentation masks. 
The object texture is controlled by the noise pixels in the input mask.
The masks used in this work are directly from the manually labelled ground-truths.
It is thus hard to synthesize objects with shape diversity.
Though both studies showed that synthetic data could be helpful for improving the performance of the supervised learning tasks, the capability of manipulating the object synthesis is still lacking in such methods.
In \cite{Chartsias2018FactorisedSegmentation}, the authors propose to obtain structured representations of medical image by training an auto-encoder network to factorize the input image into a segmentation mask and a 1D vector. Inspired by this work, we believe that manipulating the factorized image components could allow the manipulation of the synthetic image objects.

\subsection{Disentangled Image Generation}
Using the semantic-level information to guide the image synthesis or in-painting has been explored by several computer vision studies with natural images.
InfoGAN \cite{Chen2016InfoGAN:Nets} was proposed as an extension to GANs, to learn disentangled representations using mutual information in an unsupervised manner.
The $\beta$-VAE~\cite{betavae} and Factor-VAE~\cite{factorvae} were also proposed for unsupervised image disentangling.
DRGAN \cite{Tran2017DisentangledRecognition} was proposed to learn both a generative and a discriminative representation from one or multiple face images to synthesize identity-preserving faces at target poses.
In \cite{Hong2018LearningRepresentations}, the authors proposed to combine the information from both the semantic segmentation mask and the object bounding boxes to manipulate the object in-painting.
Different from most of the image synthesis methods starting with random noises, in \cite{Ma2018DisentangledGeneration}, the authors proposed a two-stage training strategy to train image synthesis networks.
The first training stage trains an auto-encoder network to obtain the embedding of the real images; the second stage maps the noise from a Gaussian distribution to the embedding distribution and then to the real data.
Similar to \cite{Zhao2018Multi-ViewSingle-View}, we use the KL divergence to train the  embedding distribution to be close to a standard normal distribution instead of training another distribution mapping.
Sun et al. and Di et al. proposed to split the face synthesis into two sub-tasks: (1) facial landmark generation from image context (2) facial landmark conditioned head in-painting \cite{Sun2017NaturalInpainting,Di2017GP-GAN:Landmarks}.

\begin{figure}
    \centering
    \includegraphics[width=1.0\linewidth]{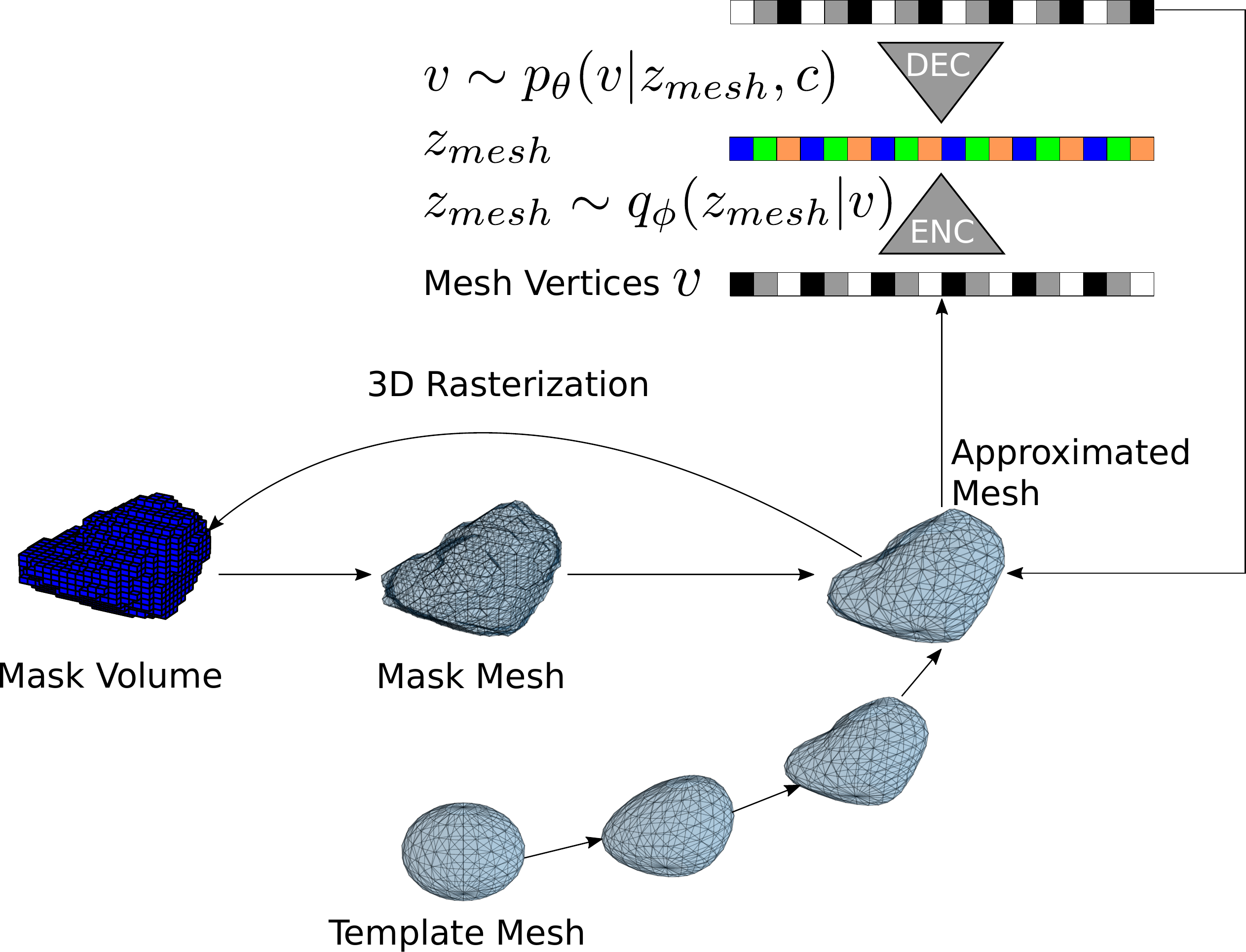}
    \caption{The shape synthesis framework with the mesh approximation and the mesh variational auto-encoder.}
    \label{fig:shape-flow}
\end{figure}

\begin{figure}[t]
    \centering
    \includegraphics[width=1.0\linewidth]{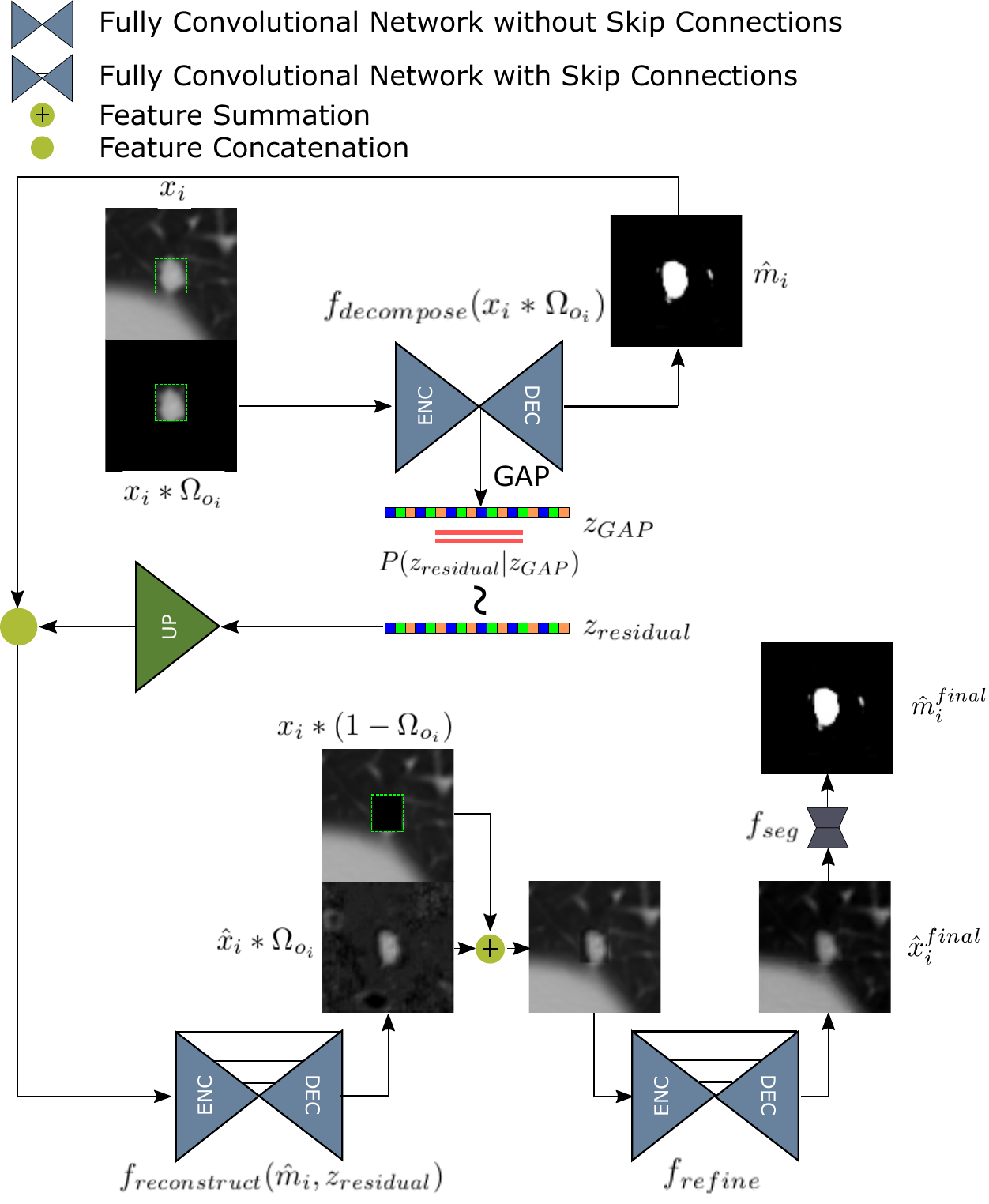}
    \caption{Stage 1: Training the image object decomposition auto-encoder network. The object area is decomposed to a segmentation mask $\hat{m}_i$ and a 1-D vector $z_{residual}$ containing the residual information to reconstruct the object. The encoder network does not have skip connections between the blocks of the same connections. 
    $z_{residual}$ is drawn from the learned distribution which is trained to be the standard normal distribution.
    The decoder firstly reconstructs the object area with the two decomposed components then blend the reconstructed object into the context with a refinement network.}
    \label{fig:decompose}
\end{figure}

\section{Methods}
We first obtain a 3D shape synthesizer by training a conditional VAE on the annotated segmentation masks. Then an auto-encoder-like network is trained to (1) decompose the object of interests into a segmentation mask and a residual embedding vector, and (2) reconstruct the object of interests from a segmentation and a residual vector. Finally, we finetune the decoder of the last stage to blend the reconstructed object into the image background originally without the presence of the target object. 

\subsection{Shape Synthesis in 3D}

\begin{figure}
    \centering
    \includegraphics[width=1.0\linewidth]{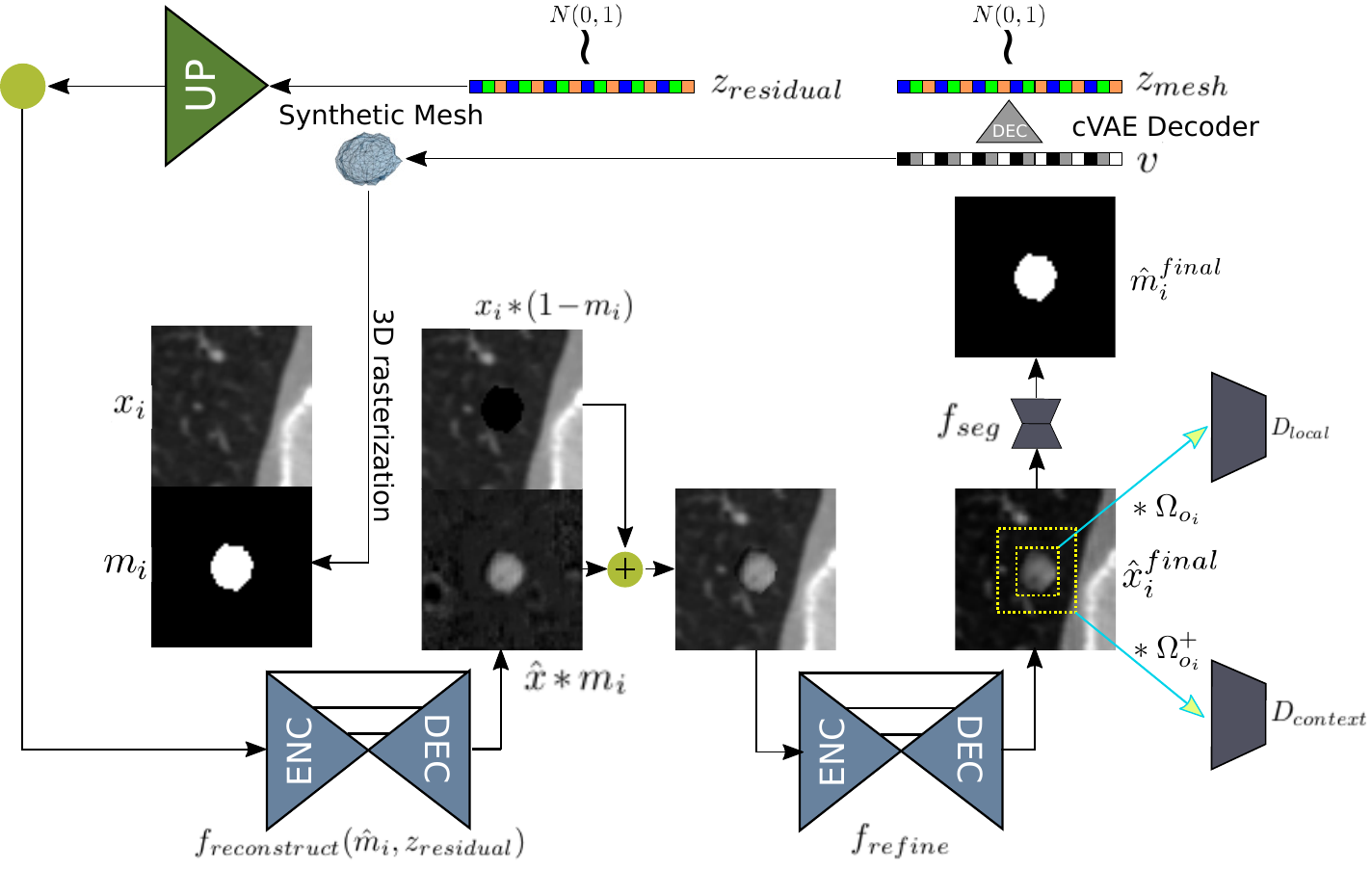}
    \caption{Stage 2: training the image synthesizer to in-paint on a random background patch originally without the presence of the object of interests. The input masks are randomly generated using the shape synthesizer. The input vector $z_{residual}$ is drawn from a standard normal distribution. Two WGAN discriminators are added to the output of the synthesizer.}
    \label{fig:synthesis}
\end{figure}

As shown in Fig.~\ref{fig:shape-flow}, we use a conditional variational autoencoder (cVAE) to obtain (1) an encoder of the object shapes that compresses the shape parameters into the distribution of a compact 1D vector, and (2) a decoder capable of reconstructing the shape given a vector randomly sampled from the standard normal distribution. We use the coordinates of the 3D mesh vertices to represent the object shape.

To approximate different nodules with a consistent parameterization, we fit a template mesh with fixed topology to binary segmentation masks. Object shapes are thereby parameterized by the mesh vertex coordinates. 
Specifically, a spherical template mesh with $N$ vertices is registered to the marching-cube-based isosurface of each mask using the coherent-point-drift algorithm \cite{CPD}.
Each 3D shape is thus represented by a 1D vector $v$ of the length $3N$. A cVAE is trained with the input $v$ as
\begin{equation}
    z_{mesh} \sim q_{\phi}(z_{mesh}|v)
\end{equation}
\begin{equation}
    v \sim p_{\theta}(v|z_{mesh},c)
\end{equation}
where $q_{\phi}(.)$ is the encoder with the weights $\phi$ that maps the shape parameters $v$ to the distribution of the latent variables; $z_{mesh}$ is an embedding vector drawn from the distribution $q_{\phi}(.)$; $\hat{v}$ is an output shape parameter vector that is reconstructed by the decoder $p_{\theta}(.)$; $c$ represents 3 conditional parameters that controls the scale and aspect ratio of the generated shape which are respectively the L2 norm of all the 3D coordinates in each dimension $\|v_x\|$, $\|v_y\|$, $\|v_z\|$. Although the scale of an object could be set analytically, our construction captures the correlations between nodule shape and size.
The cVAE is optimized by combining the L1 reconstruction loss and the KL divergence of the latent variables.
To generate a random shape, $z_{mesh}$ is sampled from the standard normal distribution as
\begin{equation}
    z_{mesh} \sim N(0, 1),\quad \hat{v} \sim p_{\theta}(v|z_{mesh},c)
\end{equation}
A result binary mask is derived from the generated mesh by 3D rasterization.

\subsection{Stage1: Image Object Decomposition}

To generate an object $o$ in an image $x$, we formulate the problem as learning an invertible distribution matching 
\begin{equation}
    z \sim q_{forward}(z|o)
\end{equation}
\begin{equation}
    o \sim p_{inverse}(o|z)
\end{equation}
where $z$ is a set of latent variables that could represent the objects of interests. To fit the generated object in a real-world image $I_i$, an additional transform is needed to blend the object into the background, making image indistinguishable to the real world images containing similar objects
\begin{equation}
    I \sim r(I|o \odot I_i)
\end{equation}
where $\odot$ defines the operation of fusing the generated object and an real-world image $I_i$.
To make part of $z$ manipulable and interpretable, $z$ can be decomposed as $z=\{z_{manipulable}, z_{residual}\}$ where $z_{manipulable}$ contains the parameters that can be specified with known properties such as the size and the intensity; $z_{residual}$ contains the residual information needed to represent the object. In this work, we decompose $z$ as the instance segmentation of the object $z_{shape}$ and a residual vector $z_{residual}$ that contains the information of the textures and boundary appearance.
Given an image patch $x_i$ and the instance segmentation of the object $m_i$, we train an auto-encoder like architecture to decompose the masked image patch $x_i * \Omega_{o_i}$ into the shape mask $\hat{m}_i$ and a residual vector $z_{residual}$ as
\begin{equation}
    \hat{m}_i, z_{residual} = f_{decompose}(x_i * \Omega_{o_i})
\end{equation}
\begin{equation}
    \hat{x}_i * \Omega_{o_i} = f_{reconstruct}(\hat{m}_i, z_{residual})
\end{equation}
where $f_{decompose}(.)$ is built with a 2D hour-glass network which outputs a binary segmentation mask $\hat{m}_i$ with the same size as the input.
We use $*$ to denote Hadamard product throughout the paper.
$\Omega_{o_i}$ is the bounding box region covering the object $o_i$.
The binary dice loss $L_{dice}$ is used to optimize the network to segment the correct masks.

By applying the global average pooling on the output features of the bottom block of $f_{decompose}$, we obtain a 1D vector $z_{GAP}$ and forward it to two fully connected layers $f_{dist}$ to output the distribution parameters of $P(z_{residual}|f_{dist}(z_{GAP}))$ where $z_{residual}$ is sampled from.
$P(z_{residual}|f_{dist}(z_{GAP}))$ gives a smooth manifold for randomly sampling $z_{residual}$ for the training stage 2 and inference.

The input of $f_{reconstruct}$ is the permuted $B\times D \times1\times1$  tensor of $z_{residual}$ with where $B$ and $D$ are respectively the batch size and the feature dimension.
$f_{reconstructs}$ progressively upsamples $z$ with upsampling layers and 2D $3\times3$ convolutional blocks with the strides $1$ till the resampled features are of the same size as $\hat{m}_i$. 
Then the upsampled features are concatenated with $\hat{m}_i$ and fed into a Res-UNet to output the masked area of the input image $\hat{x}_i * \Omega_{o_i}$ where $\Omega_{o_i}$ is a rectangle mask surrounding the object $o_i$.
The reconstructed object area is added to the background patch $\hat{x}_i * (1-\Omega_{o_i})$ to form the initial in-painting. To blend the reconstructed object into the context, the fused patch is then fed into a fully convolutional neural network $f_{refine}$ to reconstruct the entire patch in $\hat{x}_i^{final}$.
Another segmentation network $f_{seg}$ is added on top of the final reconstruction. It is optimized to segment the mask $\hat{m}^{final}_i$ from the final output $\hat{x}_i^{final}$ to reproduce $m_i$, regularizing $f_{refine}$ to preserve the original shape.
The reconstruction loss can be summarized as
\begin{equation}
L_{local}=|\hat{x}_i * \Omega_{o_i} - x_i * \Omega_{o_i}|
\end{equation}
\begin{equation}
L_{global}=|f_{refine}(\hat{x}_i * \Omega_{o_i} + {x_i} * (1-\Omega_{o_i})) - x_i|
\end{equation}
\begin{equation}
L_{dice}=\frac{2|\hat{m}_i * m_i|}{\|\hat{m}_i\|_2^2 + \|m_i\|_2^2} + \frac{2|\hat{m}_i^{final} * m_i|}{\|\hat{m}^{final}_i\|_2^2 + \|m_i\|_2^2} 
\end{equation}

\begin{equation}
    L_{recon}= \lambda_1 L_{local} + \lambda_2 L_{global} + \lambda_3 L_{dice} - \lambda_{D_{KL}} D_{KL}
    \label{eq:obj1}
\end{equation}
Here, the term $D_{KL}=D[N(\mu(x_i),\sigma(x_i)) \| N(0,1)]$ is the KL divergence that regularizes the distribution $P(z_{residual}|z_{GAP})$, so that we can sample $z_{residual}$ from a standard normal distribution $N(0,1)$.

\subsection{Stage 2: Object Synthesis on Random Patches}
After the image decomposition training, the network $f_{decompose}$ is discarded since it was used for helping the network $f_{reconstruct}$ to learn the latent embedding and a segmentation mask to an image object.
The weights of the networks $f_{reconstruct}$, $f_{refine}$ and $f_{seg}$ are preserved for finetuning the system to synthesis objects at random locations of the images.
For this training stage, we use random negative patches $x_i$ that do not contain the object of interests as the input background patches.
The trained 3D shape synthesizer is used to generate masks $m_i$ with different sizes and shapes.
The masks $m_i$ are fed into the object reconstruction network $f_{reconstruct}$ together with a random embedding vector sampled from the standard normal distribution $z_{residual}$.
The masked output $\hat{x} * m_i$ of $f_{reconstruct}$ is added to the masked background patch $x_i * (1-m_i)$ to form a coarse synthetic patch. Different from the stage of training the image decomposition, we use the synthesized mask here to mask out the background rather than using a squared mask because (1) the mask $m_i$ is more reliable at this stage (2) the final synthesized image could otherwise suffer from unnecessary artefacts at the squared mask boundaries. This patch is fed into $f_{refine}$ to blend the synthetic object into its context and obtain the final output $\hat{x}_{i}^{final}$. 
We use two Wasserstein GAN (WGAN) \cite{Arjovsky2017WassersteinGAN,Gulrajani2017ImprovedGANs} discriminators $D_{local}$ and $D_{context}$ on $\hat{x}_{i}^{final}$ to improve the appearance of the output object.
$D_{local}$ is applied to the masked area of the output patch $\hat{x}_{i}^{final} * \Omega_{o_i}$; $D_{context}$ is applied to a larger region of the output $\hat{x}_{i}^{final} * \Omega^{+}_{o_i}$ to discriminate if the synthetic object has been blended into the background as the real objects.
The weights of $f_{reconstruct}$ are frozen throughout this stage.
Both discriminators are built with a small DenseNet \cite{Huang2016DenselyNetworks} with spectral normalization \cite{Miyato2018SpectralNetworks} in each convolutional layer.
The objective function for the generator can be summarized as 
\begin{equation}
    L_{G} = w_1 L_{local} + w_2 L_{global} + w_3 L_{dice} - \lambda_D L_D
\end{equation}
where $L_{global}$, $L_{dice}$has the same definition as the terms in Eq.~\ref{eq:obj1}; the $L_{local}$ here is the L1 loss between the surrounding areas $\Omega_{s}=Dilate(m_i)$ - $m_i$ of the final reconstruction $\hat{x}_{final}$ and the corresponding areas of the original patch
\begin{equation}
L_{local}=|\hat{x}_{final} * \Omega_{s} - x_i * \Omega_{s}|
\end{equation}
$L_D$ is the weighted sum of the losses from the local discriminator and the context discriminator which are trained with the WGAN criteria
\begin{equation}
    L_{D} = \lambda_{local} L_{D_{local}} + \lambda_{context} L_{D_{context}}
\end{equation}
\begin{equation}
    \begin{multlined}
    L_{D_{local}} = E_{x_i}[D_{local}(x_i * \Omega_{o_i})]\\ - E_{z,mi}[D_{local}(\hat{x}^{local}_{i})] - \lambda_{gp} G(D_{local})
    \end{multlined}
\end{equation}

\begin{equation}
    \begin{multlined}
    L_{D_{context}} = E_{x_i}[D_{context}(x_i * \Omega^{+}_{o_i})] \\- E_{z,mi}[D_{context}(\hat{x}^{context}_i)] - \lambda_{gp} G(D_{context})
    \end{multlined}
\end{equation}
where $\hat{x}^{local}_{i}=\hat{x}^{final}_{i} * \Omega_{o_i}$; $\hat{x}^{context}_i=\hat{x}^{final}_{i} *  \Omega^{+}_{o_i}$; $G(D_{*}) = E_{\hat{x}^{final}_{i} * \Omega_*} [(\|\nabla D_{*}(\hat{x}^{*}_i)\|_2 - 1)^2]$ is the gradient penalty \cite{Gulrajani2017ImprovedGANs}.

The trained generator networks $f_{reconstruct}$ and $f_{refine}$ can be used for placing random synthetic objects $o_i$ of diameters $d_i$ at random locations $(x,y,z)$ in a 3D image volume.
Though the 3D shape synthesizer is trained by conditioning on the size to learn the correlations between the shape distribution and the object sizes, it does not guarantee that the output mask will be of the precise size as expected. We instead re-scale the generated mesh to the target size and rasterize it to a 3D mask.
We crop the 3D patch surrounding $(x,y,z)$ and feed the decomposed 2D slices to the trained $f_{reconstruct}$ and $f_{refine}$.
Before adding the output of $f_{reconstruct}$ to the masked background, we multiply it with a scale factor $[0.5, 1.5]$ to adjust the intensity of the generated object.
The 2D outputs of $f_{refine}$ are stitched into a 3D patch before being put back to to the original 3D volume.

\subsection{Application: Hard-Case Sampling of Synthetic Patches for Improving Lung Nodule Detection}
\label{sec:hard_select}
One example application of the proposed framework is to improve the performance of the pulmonary nodule detection systems.
Such systems are normally built with 2-stage coarse-to-fine network training as in ~\cite{detection_miccai}: (1) A fully convolutional neural network with a large receptive field is trained to obtain the nodule candidates; (2) A patch classifier is trained on the candidate patches to reduce the number of false positives.
When training the 3D patch classifier network, the positive patches are sampled from both the synthetic patches and the real patches in each batch. 
We control the proportion of the synthetic patches to be between 20\% to 50\%.
The hard cases in the synthetic patches can be selected based on the output of a patch classifier trained with real data only and the output of the trained discriminators.
Since the synthetic patches are all constructed to contain a nodule in it, the patches with low classifier probability are considered as hard positives.
At the same time, we would also like to only preserve the nodule patches that look real, because the knowledge learned from such patches could be generalized to the unseen data.
We use the output from the local discriminator $D_{local}$ to discard 20\% the synthetic patches with low quality from the training set.
\begin{table}[]
\begin{tabular}{@{}l|lll@{}}
\toprule
Train     & Volumes & Nodules & Negative Patches \\ \midrule
In-house        & 340     & 555     & 48595            \\
Luna      & 799     & 1100    & 56103            \\
NLST      & 1456    & 2192    & -                \\
Synthetic & -       & 47400   & -                \\
Sum       & 2595    & 51247    & 104698           \\ \midrule
Test      &  &  &                  \\ \midrule
In-house  & 205     & 300     & -  \\ \bottomrule
\end{tabular}
\caption{The breakdown of the CT images used for evaluating the nodule synthesis framework and the nodule detection. We kept the data split consistent among these two tasks.}
\label{tab:data}
\end{table}
\section{Experiments and Results}
\subsection{Data}
We acquired the chest CT images with lung nodules from the LUNA16 challenge dataset\cite{Setio2017ValidationChallenge.}, the NLST cohort \cite{NLST} and an in-house dataset.
The breakdown of the three datasets is shown in Table~\ref{tab:data}. 
We reserved the test images from our in-house dataset which were reviewed by experienced radiologists.
Because the original NLST images were only annotated with the slice number of the nodules, we had radiologists annotate the precise 3D locations of the nodules.
The NLST images were only used for extracting positive training patches since not all the nodules were guaranteed to be annotated.
We extracted positive training patches with a nodule centered in the image.
The negative training patches are sampled within the lung area without nodule appearance.
The patches are sampled with the size $64 \times 64 \times 32$ under the resolution of $0.6\times0.6\times1mm$.
The image patches are clipped with $[-1024, 600]$ Hounsfield unit (HU) values and rescaled to $[0, 255]$.
We generated the segmentation masks of the lung nodules for all the positive CT patches with a 3D DenseUNet that was trained on 710 images (LUNA \cite{Setio2017ValidationChallenge.} subset 2 to subset 9) obtained from the LIDC dataset \cite{Armato2011TheScans}.
The segmentation masks are used for both training the shape synthesizer and the image object decomposition network $f_{decompose}$.
With the trained nodule synthesizer, we synthesized 47400 3D positive nodule patches with the background patches randomly sampled from the lung area of the training images in all three datasets. To generate the synthetic masks, we randomly sampled the shape embedding from a standard normal distribution and re-scaled the synthetic meshes to make sure the diameters of the synthetic nodules are uniformly distributed between 4mm and 30mm.

\begin{figure}
    \centering
    \includegraphics[width=1.0\linewidth]{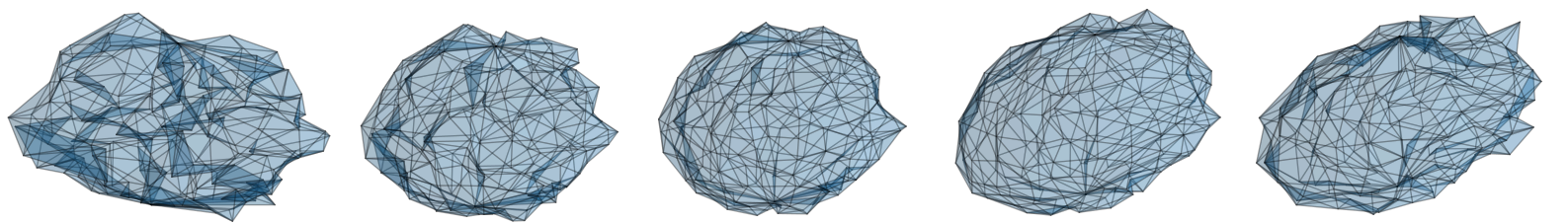}
    \caption{Example synthetic meshes of lung nodules generated from the shape synthesizer.}
    \label{fig:shapes}
\end{figure}
\subsection{Architecture and Training}
The shape synthesizer VAE is built with a multi-layer perceptron with the ReLU activation. The encoder has 3 layers which compress the input of 1452 template 3D vertices to the variational embedding of 100 variables. The decoder is built with the symmetric architecture with a linear output. This VAE directly learns the distribution of the 3D coordinates of the transformed meshes. The network was optimized using AMSGrad \cite{Reddi2018OnBeyond} with the learning rate of $1\times10^{-3}$ and the batch size of 512.

The encoder of $f_{decompose}$ is built with 3 ResNet blocks with a $2\times2$ max-pooling each and a bottom ResNet block without max-pooling.
$z_{residual}$ is obtained from the output of the bottom block with 256 feature maps. The feature maps are firstly converted into a 1D vector using the global average pooling and fed into two separate fully connected layers to obtain the variables for sampling $z_{residual}$.
The $f_{reconstruct}$ firstly uses the 6 pairs of a $2 \times 2$ upsampling layer and a $3\times3$ convolutional layer to upsample $z_{residual}$ to the original patch size. The feature maps are then concatenated with the predicted image segmentation mask and fed into a Res-UNet. $f_{refine}$ has the identical architecture as the Res-UNet in $f_{reconstruct}$.
AMSGrad \cite{Reddi2018OnBeyond} is used for optimizing all the networks used in image decomposition and refining. We use the initial learning rate of $1\times10^{-3}$ for training all the networks in the generators except the discriminators. The discriminators are trained with the initial learning rate of $1\times10^{-4}$.
To balance the GAN loss with the L1 loss in the training stage 2, we fixed $\lambda_D$ to be $0.1$.

\subsubsection{Baseline}
To compare our proposed methods with the conventional in-painting methods, we also implemented a baseline 3D in-painting method that resembles the pulmonary nodule in-painting framework proposed in \cite{Jin2018CT-RealisticSegmentation}. 
The generator network was built with a 3D Res-UNet. A WGAN discriminator was built with a 3D DenseNet.
Note that these networks are 3D networks, as it does not make sense to stitch the 2D network outputs into 3D if the generator is not trained by conditioning on the nodule shapes.
The input of the network is a 3D lung CT patch with the center area cropped out. The networks are optimized using a combined L1 loss of the local and global areas together with the WGAN adversarial loss.
Consistent to the observation in ~\cite{Jin2018CT-RealisticSegmentation}, we also found conditioning on the random vector could hamper the performance. We introduce the generation diversity by test-time dropout in the generator network.

\subsection{Qualitative Analysis of the Synthesis Networks}
\begin{figure}
    \centering
    \includegraphics[width=1.0\linewidth]{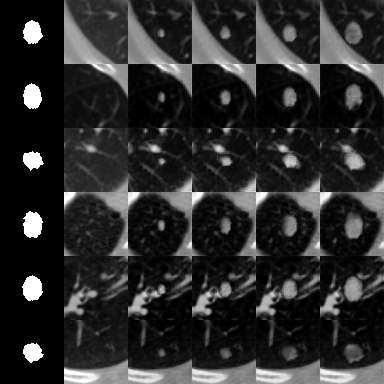}
    \caption{The synthetic lung nodules generated using the same mesh model re-scaled to different sizes. For each row, the leftmost is the central slice of the mask rasterized from a randomly generated 3D mesh. The second column is the real background patch. The rest are the generated nodule patches with increasing nodule sizes.}
    \label{fig:mask_size}
\end{figure}

\begin{figure}
    \centering
    \includegraphics[width=1.0\linewidth]{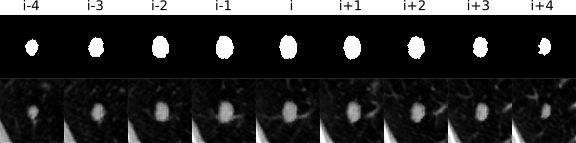}
    \caption{The different slices of a 3D generated mask (upper) and its corresponding 3D synthetic nodule patch (lower).}
    \label{fig:volume}
\end{figure}

\begin{figure}
    \centering
    \includegraphics[width=1.0\linewidth]{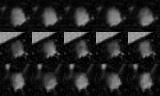}
    \caption{Each row contains the nodules generated from the same mask using different residual embedding $z_{residual}$ drawn from a standard normal distribution. $z_{residual}$ manipulates the texture within the generated nodule.}
    \label{fig:embedding}
\end{figure}

\begin{figure}
    \centering
    \includegraphics[width=1.0\linewidth]{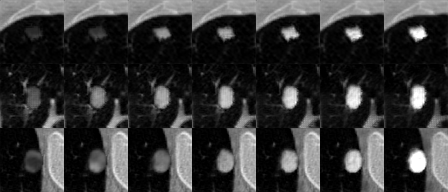}
    \caption{Each row contains the nodules generated with the identical masks and $z_{residual}$ but using an increasing scale factor between $[0.5, 1.5]$.}
    \label{fig:scale_factor}
\end{figure}

\begin{figure}
    \centering
    \includegraphics[width=1.0\linewidth]{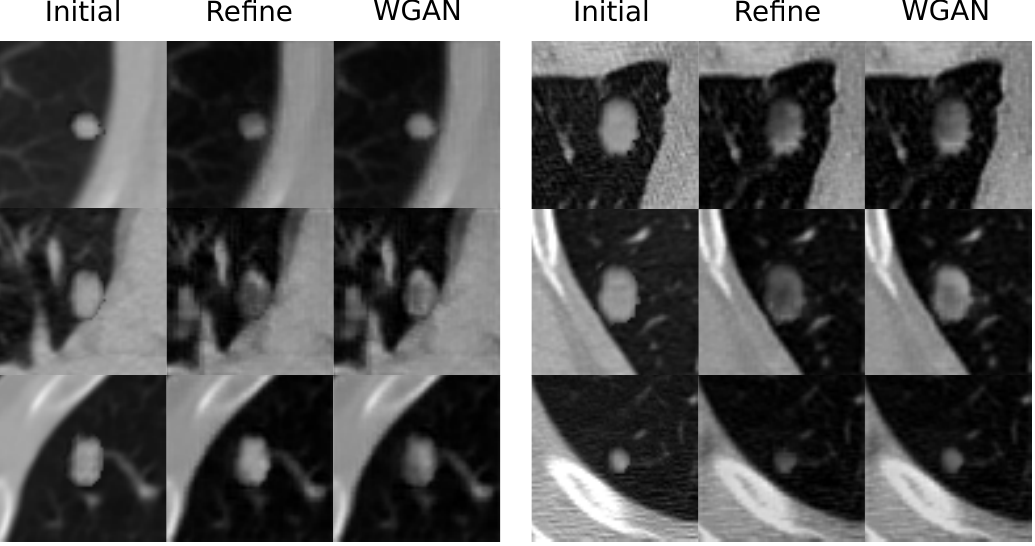}
    \caption{Six groups of synthetic nodules before the refinement network $f_{refine}$ (left), after the refine network (middle), and after the finetuning using the WGAN discriminators (right).}
    \label{fig:gan}
\end{figure}

\begin{figure}
    \centering
    \includegraphics[width=1.0\linewidth]{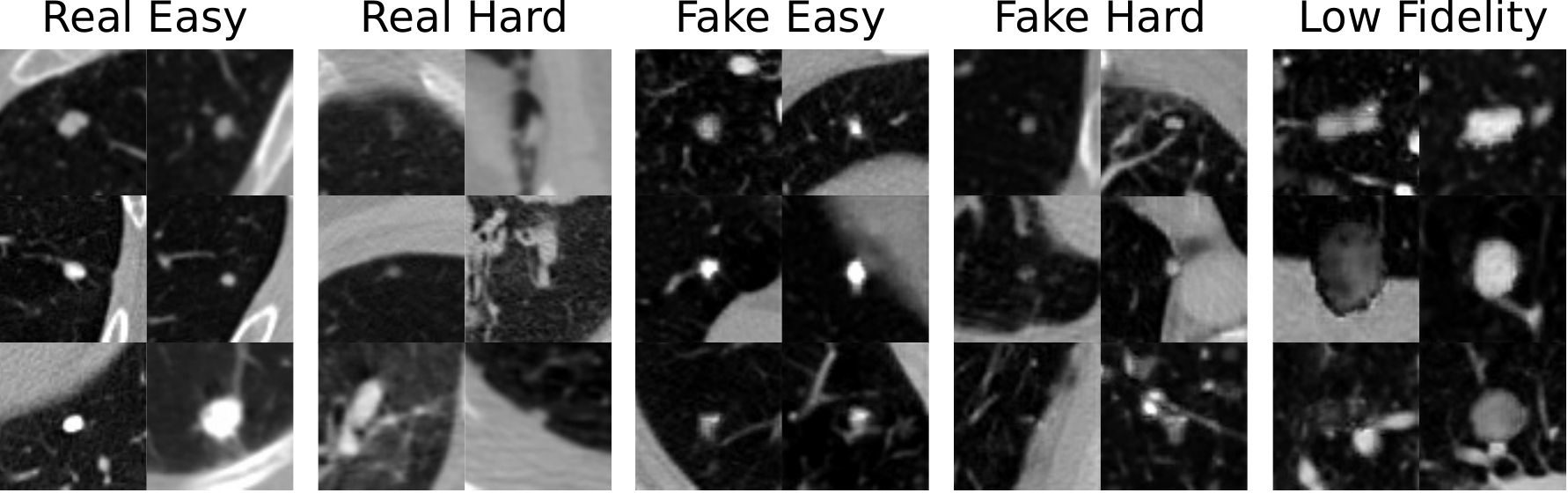}
    \caption{The example nodule patches randomly sampled from (1) real nodules with $> 95\%$ classifier output (Real Easy); (2) real nodules with $< 5\%$ classifier output (Real Hard); (3) Synthetic nodules with $>95\%$ classifier output (Fake Easy); (4) Synthetic nodules with $< 5\%$ classifier output (Fake Hard) (5) Synthetic nodules having low fidelity (lower $20\%$ of the mean local discriminator output).}
    \label{fig:hard_vs_easy}
\end{figure}

\begin{figure}
    \centering
    \includegraphics[width=1.0\linewidth]{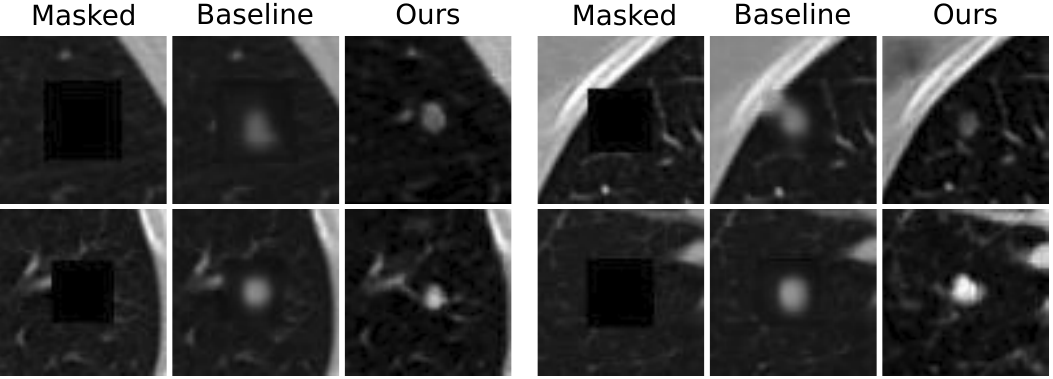}
    \caption{The visual comparison between the baseline in-painting method and our proposed.}
    \label{fig:baseline}
\end{figure}

In Fig.~\ref{fig:shapes}, we show the example shape meshes generated by the shape synthesis VAE. By sampling the hidden embedding variables from a standard normal distribution, the VAE is able to output diverse 3D meshes. Though the network was initialized randomly, most of the generated meshes tend to be roundish, resembling real pulmonary nodules.
We show in Fig.~\ref{fig:mask_size} how the same generated mesh can be re-scaled to generate lung nodules of different sizes in the image. The refine networks are able to slightly alter the appearance of the nodule to blend the generated nodule into the context.
Though the image synthesis networks were trained in 2D, we show in Fig.~\ref{fig:volume} that the nodule in contiguous slices could remain consistent to its volumetric shape.
The generated object slices are conditioned on the segmentation slices as well as its residual embedding. 
In Fig.~\ref{fig:embedding}, we show the zoomed-in synthetic nodules with the same masks and different randomly sampled residual vectors. 
The residual vectors could manipulate the textures inside the synthetic nodules as well as slightly alter the nodule boundaries.
By fixing the shapes and the residual vectors, we show that the intensity of the generated nodules can also be controlled by the intensity scale factor in Fig.~\ref{fig:scale_factor}.
In Fig.~\ref{fig:gan}, we compare the synthesis results from the network with and without the last training stage with WGAN discriminators.
The adversarial training is helpful for refining the intensities at the core and the boundaries of the nodule to blend them into the tissue context.
In Fig.~\ref{fig:hard_vs_easy}, we present example patches from the real and synthetic patches. 
We define the patch as easy when the classifier output is larger than 95\% or hard when the classifier output is smaller than 5\%.
In both the real and fake patches, the nodules with high-intensity solid cores are easier to be classified.
The hard patches tend to be of smaller sizes and low average intensity. 
It also confuses the classifier when the nodule is hidden beside the pulmonary wall or other high-intensity tissues such as the vessels or other types of abnormalities.
We define the patch with low fidelity when the mean output of the local discriminator is in the lower 20\% of the training set. It is easier for the discriminator to tell a synthetic patch contains a nodule with larger than the average diameter or irregular shape. The generator also does not handle the boundary well when it is asked to generate a large nodule besides the pulmonary wall because it is supposed to preserve the nodule boundaries of the training process. In Fig.~\ref{fig:baseline}, we compare example results (Ours) of our proposed methods with the results of the baseline method (Baseline). 

\subsection{Quantitative Analysis of the Synthesis Networks}
\begin{figure}
    \centering
    \centering
    \includegraphics[width=1\linewidth]{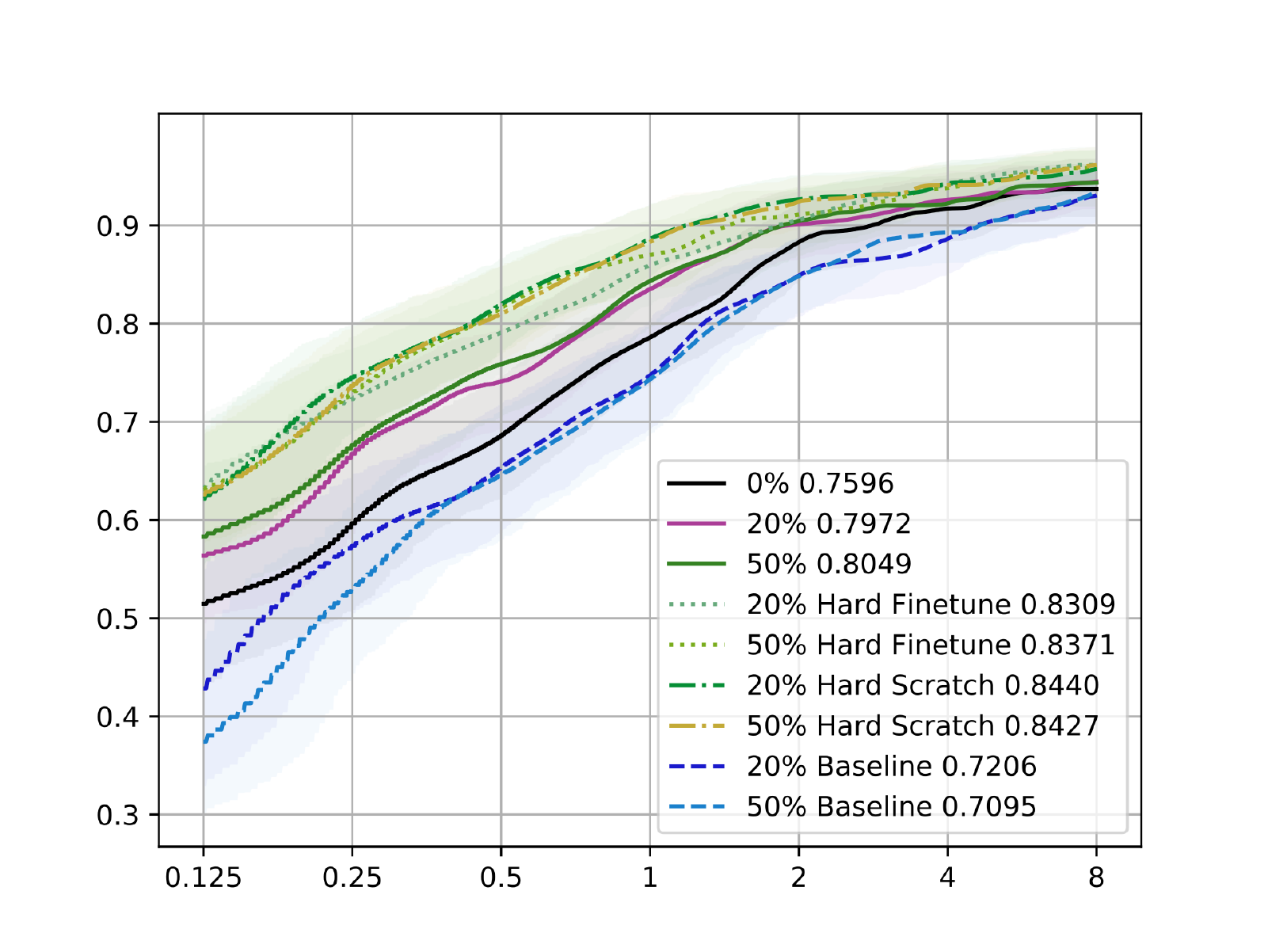}
    \caption{The nodule detection FROC curves on the $205$ test CT volumes. The x and y axis indicate the sensitivity and the number of false positive findings per volume.
    }
    \label{fig:FROC}
\end{figure}

We focus on the results of the second stage of the nodule detection framework by freezing the candidate generation network and only training the 3D patch classifier with different settings.
The patch classifier is a 3D ResNet50 with the weights pre-trained the videos in the Kinetic dataset \cite{Hara2017CanImageNet,Kay2017TheDataset}.
We applied the same set of conventional data augmentation techniques, including 90-degree rotation, random scaling and 3 direction flipping, to all the experiments for fair comparison.
In Fig.~\ref{fig:FROC}, We compare the FROC curves and the competition performance metric (CPM) scores \cite{CPM} on the test images for sampling different proportion of the synthetic patches together with all the real patches (1) training without sampling from the synthetic patches (2) with 20\% of the patches sampled from all the synthetic samples (20\%) (3) with 50\% of the patches sampled from the synthetic samples (50\%). We show that the synthetic data can be helpful for improving the detection performance especially when the number of false positives is low.
Using more than $20\%$ only slightly improve the classification performance. The confidence bands were generated with bootstrapping.
With the same sampling strategy, the patches generated by the baseline in-painting method (Baseline) did not show improvement.
In our experiment, we also tried to sample the positive patches only from the synthetic patches which did not work well because the synthetic patches do not cover the entire distribution in the real data, for example, sub-solid nodules.
We also obtained higher detection performance by only sampling from the hard cases selected based on the criteria described in S.~\ref{sec:hard_select} (Hard). 
We observed that training with the batches mixed with real and the selected hard-synthetic patches work (Scratch) slightly better than finetuning the classifier already trained on real-data only (Finetune).

\section{Conclusions}
In this paper, we proposed the manipulable object synthesis framework for generating objects in medical images.
The proposed framework is evaluated by generating synthetic lung nodules in 3D CT volumes.
By showing the qualitative results, we demonstrate that the proposed framework could synthesize realistic lung nodules at random locations with different sizes, shapes, textures, average intensities.
By evaluating on an example application of lung nodule detection using a combined dataset of CT volumes, we show that the nodules synthesized by the proposed methods could improve the overall detection performance by 8.44\% CPM score.
The detection performance can be further improved by selecting only the hard samples from the synthetic patches based on the outputs from both the patch classifier and the discriminator.
The limitations of the current framework include: (1) it does not generate high-fidelity nodules close to the pulmonary wall since the networks are trained to preserve the complete nodule shapes. This might be dealt with by constraining the nodule generation with more detailed semantic segmentation masks; and (2) the proposed simple shape synthesis methods do not support generating objects with more complex structures, for example, nested models with multiple semantic labels or connected components.

\noindent\textbf{Disclaimer}: This feature is based on research, and is not commercially available. Due to regulatory reasons, its future availability cannot be guaranteed.

{\small
\bibliographystyle{ieee}
\bibliography{manual}
}

\end{document}